\documentclass[conference]{IEEEtran}
\usepackage{cite}
\usepackage{amsmath,amssymb,amsfonts}
\usepackage{algorithmic}
\usepackage{graphicx}
\usepackage{textcomp}
\usepackage{xcolor}
\usepackage{fancyhdr}
\usepackage[hyphens]{url}
\usepackage{caption} 
\usepackage{subcaption}
\usepackage{tabularx}
\usepackage{listings}
\usepackage{adjustbox}
\usepackage{algorithm}
\usepackage{algorithmic}
\usepackage{hyperref}
\usepackage{balance}
\usepackage{footnote}
\makesavenoteenv{tabular}
\makesavenoteenv{table}

\newcommand\rev[1]{#1}

\def\BibTeX{{\rm B\kern-.05em{\sc i\kern-.025em b}\kern-.08em
    T\kern-.1667em\lower.7ex\hbox{E}\kern-.125emX}}

\title{Enabling Level-4 Autonomous Driving on a Single \$1k Off-the-Shelf Card} 
\author{
Hsin-Hsuan Sung${}^\ast$, Yuanchao Xu${}^\ast$, Jiexiong
Guan${}^\diamond$, Wei Niu${}^\circ$\\
Shaoshan Liu${}^\pm$, Bin Ren${}^\circ$, Yanzhi Wang${}^\mp$, Xipeng Shen${}^{\ast\diamond}$\\
${}^\ast$: North Carolina State University\hspace{.4in}
${}^\diamond$: COCOPLE LLC \\
${}^\circ$: College of William and Mary\hspace{.4in}
${}^\pm$: Perceptin LLC\hspace{.4in}
${}^\mp$: Northeastern Univ\\
\textit{Contact: hsung2@ncsu.edu; xipeng.shen@cocopie.ai} 
}

\begin{document}
\maketitle


\begin{abstract}
Autonomous driving is of great interest in both research and industry. The high cost has been one of the major roadblocks that slow down the development and adoption of autonomous driving in practice. This paper, for the first-time, shows that it is possible to run level-4 (i.e., fully autonomous driving) software on a single off-the-shelf card (Jetson AGX Xavier) for less than \$1k, an order of magnitude less than the state-of-the-art systems, while meeting all the requirements of latency. The success comes from the resolution of some important issues shared by existing practices through a series of measures and innovations. The study overturns the common perceptions of the computing resources required by level-4 autonomous driving, points out a promising path for the industry to lower the cost, and suggests a number of research opportunities for rethinking the architecture, software design, and optimizations of autonomous driving. 
\end{abstract}

\section{Introduction}
\label{sec:intro}

Autonomous driving is drawing great interest. The high cost has been one of the major roadblocks that slow down the development and adoption of autonomous driving in practice~\cite{liu2020autonomous}. 
One of the most costly components is the hardware to execute the autonomous driving software. 

At present, even partially autonomous (e.g., level-2) driving systems already require either a high-end accelerator box (e.g., NVIDIA Drive~\cite{nvidiadrive_2021} or some kind of custom hardware, and the cost of either is no lower than \$10k. A device that can run fully autonomous (level-4) driving software costs several times more for the much more computing resources it relies on for real-time response of the software of a larger scale and greater complexity. The Baidu autonomous driving platform, for instance, consists of two compute boxes, costs as much as \$30K, and draws 3000W power~\cite{liu2020critical}.


This paper strives to answer three research questions (RQ) crucial for the cost and hence the future development of the autonomous driving industry: 
\begin{itemize}
    \item RQ-1: Does fully-autonomous driving really need that much computing resource? 
    \item RQ-2: What causes the currently observed performance deficiency of low-end devices for fully-autonomous driving? 
    \item RQ-3: Is it possible for level-4 autonomous driving to achieve real-time performance on a single off-the-shelf card for as little as \$1K, an order of magnitude less than state of the art? How to achieve that?
\end{itemize}

To answer these questions, we conduct a focused study, trying to optimize the deployment of six level-4 autonomous driving applications (derived from Autoware~\cite{kato2018autoware}) on a single off-the-shelf low-end card, Jetson AGX Xavier~\cite{jetson_AGX_Xavier} from NVIDIA. The result overturns some common perceptions held by the industry. For the first time, we show that it is possible to run industrial-level level-4 autonomous driving on a single off-the-shelf card (Jetson) for as little as \$1k while meeting all latency requirements. Meanwhile, this study produces a set of key insights on the important pitfalls or technical deficits in the current autonomous driving industry practice and contributes several practical solutions:

\begin{itemize}
\item Deficit I: Starvation happens when prior scheduling schemes are applied to autonomous driving applications that are deployed to a single low-end device. \\
- Resolution: We propose a simple solution, {\em just-in-time priority adjustment}, which resolves the starvation by adjusting the affinity and priorities of tasks in a just-in-time manner. 
\item Deficit II: Some types of accelerators are left substantially under-utilized due to hardware-oblivious model designs and implementations.\\
- Resolution: We employ {\em hardware-aware model customization}, an approach that significantly increases the accelerators' utilization by bridging the gap between DNN models and multiple types of accelerators. 
\item Deficit III: Current scheduling algorithms for autonomous driving cannot deal with hybrid workloads that can employ multiple types of accelerators. \\ 
- Resolution: We propose {\em DAG instantiation based scheduling}, an approach that extends the scheduling of autonomous driving to meet the needs via accelerator-based DAG instantiation.
\end{itemize}

The explorations together lead to the success of making all of the six level-4 autonomous driving applications achieve real-time performance on a single Jetson card. The success has multi-fold implications. It entails the need for the industry and the research community to reexamine some assumptions (on architecture, power budget, cost, the impact of interference, etc.) commonly held on autonomous driving systems, which in turn may lead to a series of new research opportunities, such as (i) reexamining the entire system design in the backdrop of the completely different power budget and space budget, (ii) adding redundancy and reliability to low-end devices in a cost-effective manner for level-4 autonomous driving, (iii) reconsidering the research on fine-grained scheduling optimizations under the new deployment settings, (iv) reexamining the design, optimization, and deployment of other kinds of autonomous driving applications (e.g., those based on strongly-integrated multi-task DNNs).



This work, in addition, contributes an open-source research kit named {\em Single-Card Autonomous Driving Research Kit (SCAD)}. It allows the reproduction of the results in this study, easy deployment of autonomous driving on a Jetson card, as well as the generation of various autonomous driving Directed Acyclic Graphs (DAGs) for experiments and benchmarking, offering a vehicle for the community to more quickly advance the research in this field.

Research in autonomous driving is an active field, with many papers published on important research points, scheduling layers of DNNs~\cite{bateni2020neuos,bateni2018apnet,bateni2018predjoule}, scheduling memory allocation of DNNs~\cite{bateni2019predictable}, applying real-time scheduling~\cite{saito2018rosch}, supported with micro-services~\cite{tang2020container}, heterogeneity study~\cite{tang2020lopecs, liu2021pi}, and so on. This study builds on the many prior research efforts, but has a very different objective and level of focus. Rather than coming up with a better solution to a research point, this work aims to understand the whole application's performance deficits, the reasons, and practical solutions. 

It is worth noting that besides performance, accuracy, and energy efficiency, there are other aspects (e.g., redundancy and security) in our product to meet the full industry standard. These support increases the cost but only modestly, as Section~\ref{sec:discuss} discusses.

\begin{figure}
    \centering
    \includegraphics[width=0.48\textwidth]{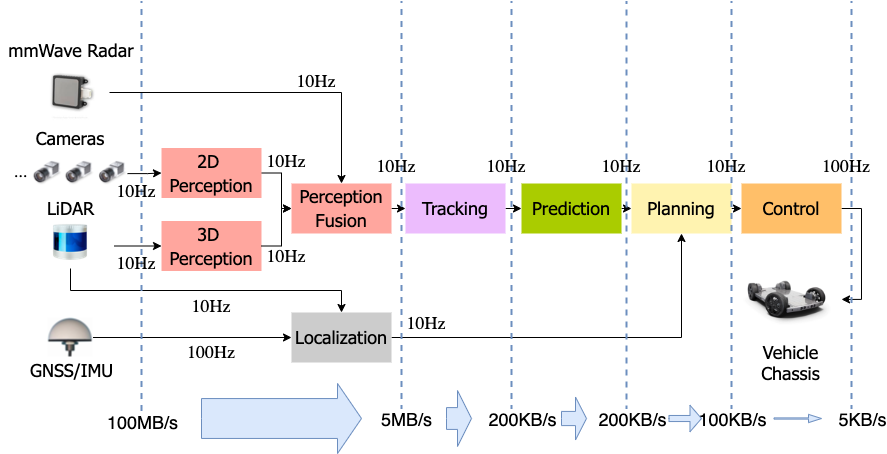}
    \caption{High-level workflow of a level-4 autonomous vehicle.}\label{fig:am}
\end{figure}

\begin{figure}
    \centering
    \includegraphics[width=0.48 \textwidth]{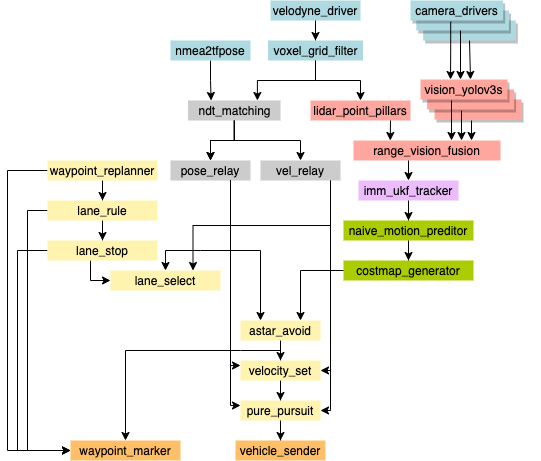}
    \caption{Detailed tasks within an example autonomous driving application (ADApp) derived from Autoware. We categorize the tasks into the modules in Figure~\ref{fig:am}: Sensing (blue), Perception (red), Localization (gray), Tracking (purple), Prediction (green), Planning (yellow), and final control output (orange)} \label{fig:DAG_Autoware}
    
\end{figure}

\section{Autonomous Driving Application and Device} 
\label{SectionIII}

In industry, level-4 is considered to be fully autonomous driving, which is of the primary interest of the current industrial research. This section explains the high-level workflow and presents an example level-4 autonomous driving application, which will be used in the following discussions.

\subsection{High-level Workflow} 
\label{sec:highLevel}

Figure~\ref{fig:am} illustrates the high-level structure of a level-4 autonomous vehicle's workflow. (i) Starting from the left side, the sensing system generates raw sensing data from mmWave radars, LiDARs, cameras, Global Navigation Satellite System (GNSS) receivers, and Inertial Measurement Units (IMUs), where each sensor produces raw data at its own frequency. For instance, the cameras capture images at 30 FPS, the LiDARs capture point clouds at 10 FPS, the GNSS/IMUs generate positional updates at 100 Hz. Certain temporal and spatial synchronizations~\cite{liu2021sync} are used. To collect enough info, a vehicle is often equipped with multiple sensing devices (e.g., cameras) of a type. (ii) The perception components create a comprehensive perception list of all detected objects. The perception and localization system consumes 100 MB/s of raw sensing data and produces 5 MB/s of semantic data in real-time. (iii) The perception list is then fed into the Tracking node at 10 Hz to create a tracking list of all detected objects. (iv) The tracking list is then fed into the Prediction node at 10 Hz to create a prediction list of all objects. The Tracking and Prediction system consumes 5 MB/s of perception inputs and further reduces the data size to 200 KB/s. (v) At last, both the prediction results and the localization results are fed into the Planning node at 10 Hz to generate a navigation plan~\cite{liu2020creating}. (vi) The navigation plan is then fed into the Control node at 10 Hz to generate control commands, which are finally sent to the autonomous machine for execution at 100 Hz. 

In every 10 ms, the autonomous machine needs to generate a control command. If any upstream node, such as the Perception node, misses the deadline for generating output, the Control node must still generate a command, which would be based on out-of-date info and hence potentially lead to disastrous results as the autonomous machine would then be essentially driving blindly without timely participation from the perception unit. Real-time response is hence essential for all the main components in an autonomous driving application.

\begin{table}[h]
    \centering
        \caption{Differences among the ADApp applications}
    \label{tab:ADAppSet}
    \begin{tabular}{|c|l|l|p{3.8cm}|}\hline
    Application     &  \multicolumn{3}{|c|}{2-D Model} \\ \cline{2-4}
    name            & model name & num & description \\\hline
    ADy288     & Yolo-v3-288 & 10 & ten Yolo-V3 models processing ten streams of videos of $288\times288$ resolution each frame  \\\hline
    ADy416 & Yolo-v3-416 & 5 & five Yolo-V3 models processing five streams of videos of $416\times416$ resolution each frame \\\hline
    ADy608 & Yolo-v3-608 & 3 & three Yolo-V3 models processing three streams of videos of $608\times608$ resolution each frame \\\hline
    ADs288 & SPP-v3-288 & 10 & ten SPP-V3 models processing ten streams of videos of $288\times288$ resolution each frame \\\hline
    ADs416 & SPP-v3-416 & 5 & five SPP-V3 models processing five streams of videos of $416\times416$ resolution each frame  \\\hline
    ADs608 & SPP-v3-416 & 3 & three SPP-V3 models processing three streams of videos of $608\times608$ resolution each frame \\\hline
    \end{tabular}
\end{table}

\subsection{ADApp: A Set of Level-4 Autonomous Driving Applications}

This study is conducted on ADApp, a set of six autonomous driving applications built on 
Autoware~\cite{the_autoware_foundation}, 
the most commonly used open-source level-4 autonomous driving software by the non-profit Autoware Foundation, created for synergies between corporate development and academic research to enable autonomous driving technology. \rev{It is worth noting that in our product, significant extensions have been made to Autoware to meet the industrial Level-4 requirements, including multiple input image sensing streams missing in the default Autoware workload, the added support of AI accelerators, and the use of real-time scheduling.}

The DAG in Figure~\ref{fig:DAG_Autoware} shows the backbone architecture of Autoware-based applications, a concretization of the main functions showed in Figure~\ref{fig:am}. The architecture is representative of level-4 autonomous driving, \rev{similar to the DAGs in other industrial autonomous driving systems (e.g., DiDi\footnote{https://blogs.nvidia.com/blog/2021/05/17/didi-nvidia-drive-self-driving-robotaxis/}).} Each node in the DAG is a {\em task}, corresponding to a process in the underlying middleware Robot Operating System (ROS) on which the application executes. There are 28 tasks in the application, forming complicated producer-consumer relations, represented by the edges. The colors of the tasks indicate the corresponding function blocks in Figure~\ref{fig:am}.


The right half of Figure~\ref{fig:DAG_Autoware} shows that the application gets 2D images from a number of cameras, and 3D points cloud from a LiDAR scanner. The \emph{velodyne\_driver} handles messages from the LiDAR device and down-samples them with \emph{voxel\_grid\_filter}. The node {\em lidar\_point\_pillars} detects 3D objects with a PointPillars DNN model~\cite{lang2019pointpillars}, and 2D objects with a number of YOLOv3\footnote{YOLOv3 is used because the more recent model YOLOv4 is not supported by Autoware by default; our separate tests show similar observations when YOLOv4 is used.} DNN models~\cite{redmon2018yolov3} (one for each image stream). Both are the most commonly used models in autonomous driving. The node \emph{range\_vision\_fusion} collects the detected objects from multiple DNNs and outputs the fused results. The node \emph{imm\_ukf\_tracker} uses the IMM-UKF-PDA tracking algorithm~\cite{IMM_UKF_PDA_tracker} which utilizes three combined Bayesian filters to simultaneously tackle association uncertainties, motion uncertainties and estimate non-linear stochastic motion model in real-time. The node \emph{native\_motion\_predictor} generates predicted angles and velocities~\cite{kato2018autoware} from which the node \emph{costmap\_generator} creates an obstacle map to show the passable probability around the car.

The left half of Figure~\ref{fig:DAG_Autoware} shows that the application collects the location info (in nmea messages) from Global Navigation Satellite System (GNNS), processed by the node \emph{nmea2tfpose} and then handled by the \emph{ndt\_matching} node to get the position of the vehicle on the map and its velocity with the Normal Distribution Transform (NDT) algorithm~\cite{NDT_Matching}. The info is then forwarded to other nodes by the \emph{pose\_relay} and \emph{vel\_relay} respectively. The application loads and processes the given global \emph{waypoints} (from pre-loaded 3D map) (\emph{waypoint\_replanner}, \emph{lane\_rule}, \emph{lane\_stop}). Then, without considering surrounding obstacles, \emph{lane\_select} generates a selected lane based on the given map, current position, and current velocity. It then generates local feasible \emph{waypoints} (i.e., trajectories) by \emph{astar\_avoid} with $A^*$~\cite{A_star_predition} and hybrid-state $A^*$~\cite{hybrid_A_star_predition} graph-searching algorithms. The node \emph{velocity\_set} then calculates the velocity to set, and the node \emph{pure\_pursuit} calculates the actuation commands with 
the Pure Pursuit algorithm~\cite{pure_pursuit_algo}.

The architecture is the de facto architecture in level-4 autonomous driving systems. It is hence shared by the six applications in ADApp. The differences among the six applications are in the 2-D perception component: Perception is the performance bottleneck of autonomous driving software; variations in other components may affect the recommended actions to take for the vehicle, but have no observable effect on the end-to-end performance of the software. Performance is the focus of this study. In perception, as 3-D perception devices (lidar) are much more expensive than 2-D perception devices (camera), common vehicles are equipped with only one lidar but differ in the number of cameras. The design of the ADApp set reflects the variations in the real world. 

Table~\ref{tab:ADAppSet} lists the six applications in ADApp and their 2D perception components. The first three use Yolov3~\cite{redmon2018yolov3} as the base for 2D perception; they differ in the input video resolutions. The lower the resolution is, the smaller the Yolov3 model is, and the lower the accuracy (and also time and device cost) is. To compensate for the lower accuracy, the vehicle is often equipped with more cameras, which explains the different numbers of 2D models used in the three applications. The bottom three applications use SSP~\cite{yolov3spp}. SPP is another commonly used 2D perception model. As a variation of Yolov3, it gets the best features in Max-Pooling layers. The settings of these three applications resemble the first three otherwise. These six applications are comparable to many level-4 autonomous driving systems in the industry~\cite{waymodriver,baiduapollo}. To make the results easy to reproduce, the used Yolov3 and SPP are both in the standard form; no custom compression or other changes are made to them. 

The inputs used for all the experiments are from Autoware-AI~\cite{ROSBAG_DEMO}. Following the common practice, each experiment first loads the static information (e.g., car configuration, 3D map, and lane information) before execution, and uses the recorded ROSBAG to playback the dynamic data (e.g., point clouds) to simulate the practical computing situation. Additionally, we add the artifact image message flows into the Autoware system to simulate the situations that have one LiDAR and multiple cameras.

\begin{table*}[ht!]
\renewcommand{\arraystretch}{1.2}
    \centering
    \caption{Configurations of the Target Device (Jetson AGX Xavier)}
    \label{tab:devices}
    \begin{tabular}{|c|c|c|c|c|c|c|c|c|} \hline
    Hardware     & $^{1}$ CPUs & $^{2}$ GPUs & GPU Throughput& GPU Power & DLAs& DLA Throughput & DLA Power & Cost (USD) \\ \cline{2-9}
     & 1 & 1 (1377 MHz) & 2.8 TFLOPS FP16& 30W & 2 (1395.2 MHz) & 2.5 TFLOPS FP16& 0.5-1.5W & \$699 \\ \hline\hline
     Software & \multicolumn{8}{|l|}{Jetpack 4.4.1 SDK (L4T 32.4.4 OS), CUDA 10.2, OpenCV 3.4.3 compiled with CUDA, TensorRT 7.1.3.0, cuDANN 8.0.0} \\ \hline
    \end{tabular}
     $^{1}$ 8-core NVIDIA Carmel Arm v8.2 64-bit CPU 8MB L2 + 4MB L3\\  
     $^{2}$ NVIDIA Volta architecture with 512 NVIDIA CUDA cores and 64 Tensor cores.
\end{table*}

\subsection{Device}

In the current industry, an autonomous driving software as complicated as any of the six applications in ADApp relies on high-end autonomous driving devices. For instance, according to a previous literature~\cite{liu2017computer}, the platform from a leading level-4 autonomous driving company consists of two compute boxes, each equipped with an Intel Xeon E5 processor and four to eight Nvidia GPU accelerators, connected with a PCI-E bus. The whole system costs over \$30K and at its peak consumes over 3000W power, making the whole solution unaffordable to average consumers.

Our targeted device is the off-the-shelf Jetson Xavier SoC. Table~\ref{tab:devices} lists the hardware and software configurations. The main computing units include one 8-core Carmel CPU based on ARM v8 ISA, one NVIDIA Volta-class integrated GPU, and two Deep Learning Accelerators (DLA). The integrated Volta GPU has 20 TOPS for INT8 operations and 2.8 TFLOPS for FP16 operations. DLA is a specialized accelerator for Deep Learning inference in cost-sensitive applications. It focuses on executing four common DNN operations: convolutions, activations, pooling, and normalization with power efficiency. Although it has 10 TOPS for INT8 operations that is half lower than the integrated Volta GPU, its power consumption is much lower than the integrated Volta GPU. \rev{Jetson is also equipped with a Programmable Vision Accelerator, useful for image preprocessing. However, as many products in the autonomous driving industry, our autonomous driving system deploys preprocessing on sensors themselves, not central computing devices. It is hence not considered in this exploration.} The memory on the device is 256-Bit LPDDR4 Memory with a bandwidth 136 GB/s. The overall cost of this Jetson Xavier SoC on the market is \$699. 

The software configurations include Jetpack 4.4.1 SDK that runs L4T 32.4.4 OS. It is equipped with CUDA 10.2, OpenCV 3.4.3 compiled with CUDA, TensorRT 7.1.3.0, and cuDNN 8.0.0.

\begin{table*}[!ht]
\centering
\caption{Execution time (mean$\pm$ std)  of each module in the ADApp applications on Jetson AGX Xavier and the miss rates. {\small The $\infty$ represents timeout. The \texttt{miss rate} of a module is how often the module misses its expected latency (shown in the parentheses in the table header)---up to 10\% over is allowed to tolerate system noises. The column \texttt{Miss Rate} shows the miss rates of the most sluggish modules (whose times are prefixed with an $\ast$), that is, the modules with the largest miss rate in the application. 
}}\label{tab:bigtable}
\renewcommand{\arraystretch}{1.2}
\small
\begin{tabular}{|l||c|cc|c|c|c|c||c|}\hline
Application  & \multicolumn{7}{|c||}{Running Time of Each Module (ms) \textbf{[expected latency in brakets]}} & {\bf Miss Rate} \\ \cline{2-8}
& \emph{Sensing} & \emph{3D Percept} & \emph{2D Percept} & \emph{Localization} & \emph{Tracking} & \emph{Prediction} & \emph{Planning} &  \\ 
& [100ms] & [100ms] & [100ms] & [100ms] & [100ms] & [100ms] & [10ms] & \\ \hline
\multicolumn{9}{|l|}{1. Default Linux Time Sharing (\autoref{sec:default})} \\ \hline
ADy288& $14.3\pm5.2$ & $94.7\pm12.8$ & * $193.3\pm17.5$ & $89.5\pm30.5$ & $0.9\pm0.8$ & $0.4\pm1.0$ & $1.0\pm0.1$ & $100\%$ \\
ADy416 & $15.3\pm5.1$ & $90.2\pm12.0$ & * $167.6\pm12.7$ & $89.1\pm29.1$  & $0.9\pm0.7$  & $0.5\pm0.9$ & $1.1\pm0.2$  & $100\%$  \\
ADy08 & $14.8\pm4.8$ & $89.0\pm18.7$ & * $192.8\pm16.2$  & $91.5\pm31.2$ & $1.1\pm0.7$ & $0.4\pm0.9$ & $1.1\pm0.2$ & $100\%$ \\
ADs288 & $14.3\pm5.0$ & $95.6\pm13.7$ & * $195.2\pm18.2$  & $88.7\pm28.9$ & $1.0\pm0.8$ & $0.4\pm1.0$ & $1.1\pm0.1$ & $100\%$ \\
ADs416 & $14.8\pm4.8$ & $91.3\pm13.4$ & * $168.8\pm13.3$ & $90.1\pm30.2$ & $1.1\pm0.9$ & $0.5\pm1.0$ & $1.1\pm0.0$ & $100\%$ \\
ADs608 & $14.7\pm4.9$ & $90.6\pm19.2$ & * $194.2\pm17.7$ & $91.2\pm29.3$ & $0.9\pm0.6$ & $0.4\pm1.1$ & $1.1\pm0.1$ & $100\%$ \\ \hline
\multicolumn{9}{|l|}{2. Default ROSCH (\autoref{sec:default})} \\ \hline
ADy288 & $8.8\pm1.0$ & * $\infty$ & * $\infty$ & * $\infty$ & * $\infty$ & * $\infty$ & $1.1\pm0.8$ & $100\%$\\
ADy416 & $8.5\pm0.7$ & * $\infty$ & * $\infty$ & * $\infty$ & * $\infty$ & * $\infty$ & $1.3\pm0.9$ & $100\%$ \\
ADy608 & $8.5\pm0.8$ & * $\infty$ & * $\infty$ & * $\infty$ & * $\infty$ & * $\infty$ & $1.1\pm0.7$ & $100\%$ \\
ADs288 & $9.0\pm0.8$ & * $\infty$ & * $\infty$ & * $\infty$ & * $\infty$ & * $\infty$ & $1.2\pm1.0$ & $100\%$ \\
ADs416 & $8.4\pm1.0$ & * $\infty$ & * $\infty$ & * $\infty$ & * $\infty$ & * $\infty$ & $1.3\pm0.6$ & $100\%$ \\
ADs608 & $8.5\pm0.9$ & * $\infty$ & * $\infty$ & * $\infty$ & * $\infty$ & * $\infty$ & $1.5\pm0.8$ & $100\%$ \\ \hline
\multicolumn{9}{|l|}{3. Just-In-Time (JIT) Priority Adjustment (\autoref{sec:jit})} \\ \hline
ADy288 & $8.5\pm0.9$ & $94.6\pm13.4$ & * $194.7\pm16.3$ & $43.5\pm10.2$ & $1.0\pm0.7$ & $0.6\pm1.1$ & $1.2\pm0.4$ & $100\%$ \\
ADy416 & $8.4\pm1.0$ & $91.7\pm11.2$ & * $166.8\pm11.4$ & $45.3\pm11.3$ & $0.8\pm1.0$ & $0.6\pm1.1$ & $1.0\pm0.4$ & $100\%$ \\
ADy608 & $8.7\pm0.7$ & $88.9\pm17.6$ & * $190.9\pm17.9$ & $47.2\pm9.9$ & $1.1\pm0.6$ & $0.5\pm0.9$ & $1.0\pm0.5$ & $100\%$ \\
ADs288 & $8.9\pm0.9$ & $96.1\pm12.7$ & * $195.9\pm17.3$ & $46.9\pm12.1$ & $0.9\pm0.9$ & $0.6\pm1.0$ & $1.0\pm0.4$ & $100\%$ \\
ADs416 & $9.0\pm0.7$ & $92.8\pm11.4$ & * $169.7\pm13.3$ & $48.1\pm11.9$ & $1.0\pm0.5$ & $0.5\pm1.1$ & $1.3\pm0.4$ & $100\%$ \\
ADs608 & $8.7\pm0.9$ & $91.2\pm20.3$ & * $194.8\pm16.9$ & $44.7\pm10.9$ & $0.8\pm1.0$ & $0.4\pm1.2$ & $1.2\pm0.2$ & $100\%$ \\ \hline
\multicolumn{9}{|l|}{4. JIT Adjustment + Migration to Accelerators (\autoref{sec:migratingToDLA})} \\ \hline
ADy288 & $8.7\pm0.8$ & $123.8\pm18.5$ & * $225.6\pm5.0$ & $43.0\pm9.9$ & $0.9\pm1.0$ & $0.5\pm0.9$ & $1.0\pm0.6$ & $100\%$ \\
ADy416 & $8.8\pm1.1$ & $128.7\pm12.1$ & * $177.6\pm3.3$ & $46.7\pm10.5$ & $1.0\pm0.8$ & $0.5\pm1.0$ & $1.2\pm0.3$  & $100\%$ \\
ADy608 & $9.1\pm0.9$ & $144.3\pm8.1$ & * $171.8\pm3.0$ & $48.5\pm9.8$ & $1.0\pm0.8$ & $0.6\pm1.1$ & $1.2\pm0.3$ & $100\%$ \\
ADs288 & $9.0\pm0.8$ & $125.6\pm17.1$ & * $225.6\pm6.3$ & $47.3\pm11.3$ & $1.1\pm0.7$ & $0.4\pm1.1$ & $1.3\pm0.2$ & $100\%$ \\
ADs416 & $8.8\pm1.1$ & $130.5\pm13.2$ & * $180.1\pm4.7$ & $47.6\pm9.8$ & $0.9\pm0.9$ & $0.7\pm1.2$ & $1.5\pm0.2$ & $100\%$ \\
ADs608 & $8.6\pm0.9$ & $147.2\pm9.2$ & * $174.3\pm4.5$ & $47.6\pm10.4$ & $0.9\pm0.9$ & $0.6\pm1.2$ & $1.2\pm0.5$ & $100\%$ \\ \hline
\multicolumn{9}{|l|}{5. JIT Adjustment + Migration to Accelerators + Hardware-Aware Model Customization (\autoref{sec:customization})} \\ \hline
ADy288 & $8.4\pm1.2$ & * $89.0\pm15.3$ & $95.6\pm5.1$ & $46.3\pm9.8$ & $0.9\pm0.9$ & $0.7\pm0.9$ & $1.0\pm0.4$ & $0\%$ \\
ADy416 & $9.0\pm0.9$ & $72.0\pm9.0$ & $88.1\pm4.3$ & $44.9\pm10.7$ & $1.0\pm0.8$ & $0.6\pm0.9$ & $1.3\pm0.2$  & $0\%$ \\
ADy608 & $8.8\pm1.2$ & $80.8\pm10.6$ & * $98.1\pm5.0$ & $46.4\pm11.0$ & $1.0\pm0.7$ & $0.4\pm1.1$ & $1.1\pm0.3$ & $0\%$ \\
ADs288 & $9.0\pm1.1$ & * $92.0\pm14.3$ & $96.4\pm5.4$ & $45.8\pm10.1$ & $1.1\pm0.7$ & $0.4\pm1.1$ & $1.2\pm0.6$ & $0\%$ \\
ADs416 & $8.9\pm0.8$ & $74.2\pm9.3$ & $90.0\pm4.2$ & $47.2\pm10.0$ & $1.0\pm0.8$ & $0.5\pm0.9$ & $1.2\pm0.2$ & $0\%$ \\
ADs608 & $8.8\pm1.0$ & $83.7\pm9.7$ & * $100.1\pm4.4$ & $46.8\pm9.9$ & $0.9\pm0.7$ & $0.7\pm0.8$ & $1.3\pm0.8$ & $0\%$ \\ \hline
\multicolumn{9}{|l|}{6. JIT Adjustment + Migration to Accelerators + Hardware-Aware Model Customization + Iterative Co-run Aware Scheduling (\autoref{sec:corun})} \\ \hline
ADy288 & $8.0\pm0.6$ & * $90.2\pm16.1$ & $94.7\pm5.5$ & $47.7\pm10.0$ & $1.0\pm0.7$ & $0.5\pm1.1$ & $1.3\pm0.2$ & $0\%$ \\
ADy416 & $8.5\pm0.5$ & $72.6\pm10.1$ & $87.8\pm4.5$ & $43.2\pm9.1$ & $0.9\pm0.6$ & $0.6\pm0.8$ & $1.0\pm0.1$  & $0\%$ \\
ADy608 & $8.7\pm0.4$ & $82.1\pm9.9$ & * $96.9\pm4.8$ & $44.9\pm11.9$ & $1.0\pm0.6$ & $0.5\pm0.9$ & $1.2\pm0.2$ & $0\%$ \\
ADs288 & $8.4\pm0.5$ & * $93.2\pm14.2$ & $97.0\pm5.2$ & $46.4\pm8.9$ & $0.9\pm0.7$ & $0.4\pm1.1$ & $1.0\pm0.3$ & $0\%$ \\
ADs416 & $8.3\pm0.5$ & $73.8\pm9.9$ & $91.3\pm4.7$ & $47.3\pm10.2$ & $1.0\pm0.8$ & $0.6\pm0.9$ & $1.1\pm0.8$ & $0\%$ \\
ADs608 & $8.6\pm0.4$ & $84.2\pm10.0$ & * $100.1\pm3.2$ & $46.5\pm9.7$ & $0.9\pm0.7$ & $0.5\pm0.8$ & $1.1\pm0.5$ & $0\%$ \\ \hline
\end{tabular}
\end{table*}

\begin{table*}[ht!]
    \centering
        \caption{Power consumption for the ADApp applications on Jetson AGX Xavier by INA3221 power monitors~\cite{XavierControl} (unit: milliWatts)}
    \label{tab:ADAppPower}
    \begin{tabular}{|l|c|c|c||l|c|c|c||l|c|c|c|}\hline
    \multicolumn{4}{|c||}{\autoref{sec:jit}} & \multicolumn{4}{|c||}{\autoref{sec:jit} + \autoref{sec:migratingToDLA}} & \multicolumn{4}{|c|}{\autoref{sec:jit} + \autoref{sec:migratingToDLA} + \autoref{sec:customization} } \\ \hline\hline
    Application & GPU & DLA & Total System & Application & GPU & DLA & Total System & Application & GPU & DLA & Total System  \\\hline
    ADy288 & 18,684 & 0 & 33,029 & ADy288 & 18,378 & 1,683 & 32,919 & ADy288 & 13,948 & 4,441 & 31,750 \\\hline
    ADy416 & 18,401 & 0 & 30,266 & ADy416 & 19,740 & 1,834 & 34,729 & ADy416 & 14,237 & 4,590 & 33,410 \\\hline
    ADy608 & 20,675 & 0 & 33,025 & ADy608 & 20,343 & 1,987 & 35,064 & ADy608 & 13,936 & 4,437 & 33,303 \\\hline
    ADs288 & 18,323 & 0 & 32,013 & ADy288 & 19,012 & 1,645 & 33,142 & ADy288 & 14,589 & 4,358 & 32.019 \\\hline
    ADs416 & 18,756 & 0 & 31,415 & ADy416 & 20,032 & 1,785 & 35,165 & ADy416 & 14,357 & 4,640 & 34,774 \\\hline
    ADs608 & 20,876 & 0 & 33,048 & ADy608 & 20,132 & 2,038 & 36,043 & ADy608 & 14,128 & 4,549 & 34.237 \\\hline
    \end{tabular}
\end{table*}

\section{Observations and Analysis of the Default Executions}
\label{sec:default}

In the current industry, for an autonomous driving application as complicated as those in ADApp, the system usually deploys the tasks on multiple boards equipped with many computing units.
This section reports the observations when we try to deploy each of these applications on a single Jetson AGX Xavier board in the default manner. These observations help us understand the limitations of the current deployment and resource management. 

\subsection{Observations} 
\label{sec:baseObs}

The default performance is shown in the top two segments of Table~\ref{tab:bigtable}. The first segment is about the case that uses the default Linux time-sharing scheduling (version of Linux 2.6.23)~\cite{sched_manul_linux}. The second segment is about the case that uses ROSCH~\cite{saito2018rosch}, the latest published scheduling algorithm specially designed for Autoware. (Note that in all those experiments, as well as all the other experiments reported throughout this paper, the planning module of the autonomous driving applications is bound to two CPU cores isolated from the other tasks. Such isolation is a custom in autonomous driving as it is important to ensure the planning module gives out planned actions in time.)


From the table, we can see the following:
\begin{itemize}
    \item In the Linux default case,  at every time frame, all six applications exceed the allowed latency when processing the inputs. The most sluggish module is the 2D perception, taking time about twice as much as its allowed latency. 
    \item ROSCH, the scheduling algorithm specially designed for autonomous driving, actually makes the applications perform even worse. None of the applications are able to make any progress on any input. The sensing module and the planning module are the only modules that produce outputs. The other modules are all time out. 
    (Note, as mentioned in Section~\ref{sec:highLevel}, the planning module always gives outputs in a fixed frequency; the outputs are based on obsolete information if there are delays in the earlier stages of the DAG.)
\end{itemize}

\begin{algorithm}
	\caption{Original HEFT algorithm for DAG scheduling~\cite{HEFT}}\label{fig:heft}
	\begin{algorithmic}[1]
    	\STATE Set the computation cost for each node and the communication cost for each edge in the DAG with their mean values
    	\STATE Post-traverse the DAG calculate $rank$ for each task \[ rank(n_i) = \overline{w_i} + \lim_{n_j \in succ(n_i) } (\overline{c_{i,j}} + rank(n_j))\] where the \( succ(n_i) \) is the set of immediately connected successors of task \( n_i \), \( \overline{c_{i,j}} \) is mean value of communication cost of edge \( (i,j) \) , and \( \overline{w_i} \) is average computation cost of task \( n_i \). For the task $n_{exit}$, the rank value is \[ rank(n_{exit}) = \overline{w_{exit}} \]
    	\STATE Create an ordered list (taskList) of the nodes based on the descending order of their $rank$. 
    	\WHILE {there are unscheduled tasks in the list}
            \STATE $n_i$ = next task in the taskList
    		\FOR {each processor $p_{k}$ in the processor-set $Q$}
    			\STATE Compute the earliest finish time, $EFT(n_{i},p_{k})$, with the \emph{insection-based scheduling formula}
    		\STATE Assign task $n_{i}$ to the processor $p_{j}$ that minimizes $EFT$ of task $n_{i}$.
    		\ENDFOR
    	\ENDWHILE
    	\STATE Assign priorities to the tasks assigned to the same processor based on their ranks.
	\end{algorithmic}
\end{algorithm}

\subsection{Analysis}
\label{sec:defaultAn}

The executions under the default Linux time-sharing schedule are no surprise: The scheduling scheme is not designed for real-time autonomous driving workload; its unsatisfying performance has been the motivation for the prior efforts for designing specialized scheduling algorithms for such workloads. The much worse results from ROSCH, an algorithm customized to autonomous driving, was a surprise. 

Our detailed analysis shows that a reason is starvation caused by the scheduler. The scheduling algorithm in ROSCH is a variation of the most popular real-time scheduling algorithm Heterogeneous Earliest-Finish-Time (HEFT). Figure~\ref{fig:heft} shows the algorithm. This algorithm allows the computing units to have different speeds, but assumes a node in the DAG is a scheduling unit: The entire node is assigned to a processor. At a high-level, the algorithm has two steps, ranking the tasks based on their rank metric, and then scheduling the tasks in their ranking. It assigns a task to the processor that minimizes the estimated earliest finish time (EFT) of that task. If a computing unit is assigned multiple tasks, the algorithm assigns priorities to those tasks based on their ranks. The scheduling algorithm used in ROSCH is the same as HEFT except that they use different metrics for task ranking.

When this algorithm is applied to the autonomous driving applications on a single Jetson AGX Xavier card, starvation happens. The reason is as follows. Because there are only six CPU cores left after the reservation of two cores for the Planning module, many DAG nodes are scheduled to the same core. Some tasks in the Perception module are assigned to the same core as the tasks in the Sensing module. Based on the scheduling algorithm, the ROSCH scheduler gives higher priorities to the Sensing nodes than to the Perception nodes. 
Each node is a process consisting of multiple threads. All these threads inherit the priority of the process. As a result, the threads of the Sensing node, for their higher priorities, hog the CPU core throughout the execution. The Perception nodes on the same core are starved. As the other modules depend on the output of the Perception module, they make no progress either. The exception is the Planning module. As mentioned before, the module always gives outputs in a fixed frequency; as it receives no updates from the other modules, its outputs are out of date and useless. Such a problem does not show up in  high-end devices where all main modules can get their dedicated computing units.

\section{Just-In-Time Affinity and Priority Adjustment} 
\label{sec:jit}

To address the starvation problem, we change the scheduling algorithm such that the core affinity and priority adjustments are made in a just-in-time manner. Before explaining the improved scheduling scheme, it is necessary to briefly review the background of Linux scheduling systems.

\subsection{Background on Linux Schedule}

A task in Linux may be put into one of two scheduling queues: SCHED\_FIFO and SCHED\_OTHER~\cite{sched_manul_linux}. SCHED\_FIFO is first in first out. It can only be applied to threads with static priorities above 0. When SCHED\_FIFO thread becomes runnable, it will immediately preempt any currently running SCHED\_OTHER threads. SCHED\_OTHER is used in the default Linux time-sharing scheduling. It is only applicable to threads at static priority 0 that do not require special real-time mechanisms. 

\subsection{Just-In-Time Adjustment}

In ROSCH, every task is put into a SCHED\_FIFO queue and has a statically assigned priority as calculated by the HEFT algorithm. In our improved algorithm, every task is put into a SCHED\_OTHER queue and has the same default priority. Changes are made when a work item is put into the input queue of a task, and the main thread of that task is about to be invoked to process that item. At that moment, the main thread of that task is set to SCHED\_FIFO, and its priority is changed to $p$, the level calculated by the default HEFT algorithm. Please note, the changes are made to only the main thread, and the other threads of that task keep the default priority. As soon as the main thread finishes processing the item, it is set back to SCHED\_Other, and its priority is reset to the default. 

Listing~\ref{sub_task_warp_listing} shows the implementation. On Lines 1 and 2, \texttt{core\_id} and \texttt{priority} hold the CPU core and priority that the default HEFT algorithm assigns to the node that this current main thread belongs to. Lines 10 to 15 bind the thread to the CPU core and put the thread into the SCHED\_FIFO queue with priority. The thread is now being put into real-time mode to do its work. After the work is done, lines 22 to 28 unset the core affinity and puts the thread back into the SCHED\_OTHER queue with the default priority. 

In this JIT scheme, the time for a node to hold high priority is reduced, and also the non-critical assistant threads are not promoted to a high priority, which also reduces the core contention. 


\begin{lstlisting}
\begin{minted}
[
frame=lines,
framesep=2mm,
baselinestretch=1.2,
numbersep=0pt,
fontsize=\footnotesize,
linenos
]{cpp}
    /*
    Codelet to run after the thread is triggered
    */
    
    int core_id = 1; // the core to bind 
    int priority = 98; // the priority to set
    pthread_t this_thread = pthread_self(); 
    cpu_set_t cpuset;
    CPU_SET(core_id, &cpuset);

    // set affinity and priority for real-time run
    pthread_setaffinity_np(
        this_thread, sizeof(cpu_set_t), &cpuset);
    struct sched_param params;
    params.sched_priority = priority;
    pthread_setschedparam(
        this_thread, SCHED_FIFO, &params);
    
    /* ----- main work ----- */
    ... ...
    /* --------------------- */
    
    // reset the affinity and priority 
    // to the default
    for (int i = 0; i < USABLE_CORES; i++)
        CPU_SET(i, &cpuset);
    pthread_setaffinity_np(
        this_thread, sizeof(cpu_set_t), &cpuset)
    params.sched_priority = 0;
    pthread_setschedparam(
        this_thread, SCHED_OTHER, &params);
\end{minted}
\caption{Implementation of the Just-In-Time priority adjustment}
\label{sub_task_warp_listing}
\end{lstlisting}


\subsection{Performance and Analysis}
\label{sec:jitResult}

Segment three in Table~\ref{tab:bigtable} reports the performance of the six applications after the improved scheduling scheme is applied to ROSCH. 
\begin{itemize}
    \item We can see that the starvation issue is gone. Every module makes progress. 
    \item Compared to Default Linux Time-Sharing performance, the Sensing and Localization modules show significant speedups because the real-time priority helps them get the resource when their main thread needs to do the work. 
    \item Compared to Default Linux Time-Sharing performance, the 2D perception and 3D perception, however, do not show speedups. Detailed analysis shows that it is because these tasks involve both CPU and GPU code and the GPU code takes most of the time. So even though starvation is avoided, the six applications still miss the deadline in 100\% of the time. 
\end{itemize}


\section{Migration to Take Advantage of All Accelerators}
\label{sec:migratingToDLA}

To understand how effectively the applications take advantage of the accelerators on Jetson AGX Xavier, we use the NVIDIA NSight to profile the executions of the applications (after the JIT improvement is made to ROSCH). The result indicates that the applications use the GPU extensively, but leave the two DLAs unused at all. 



\subsection{Changes}

DLAs are special deep learning accelerators designed to accelerate some common operations in DNNs. For a DNN to take advantage of DLAs, however, the applications must be written with TensorRT in some required form. 
TensorRT~\cite{tensorrtAbs} is a library developed by NVIDIA for faster DNN inference on NVIDIA devices. 


\begin{figure}
    \centering
    \includegraphics[width=0.48\textwidth]{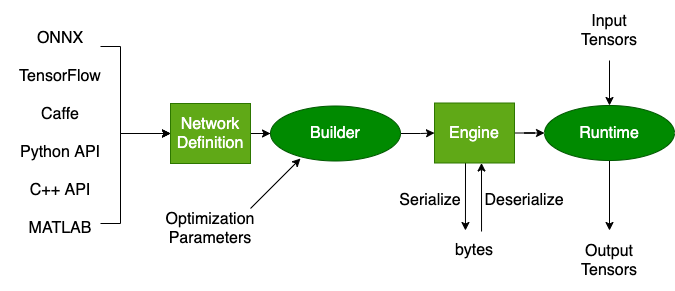}
    \caption{The TensorRT Workflow.~\cite{tensorrtAbs}}\label{fig:tensorRT}
\end{figure}

The default Autoware, despite being officially migrated to AGX Xavier, does not contain the use of the right TensorRT APIs to make use of the DLAs. As a result, the six applications in ADApp do not have those APIs used either. The lack of usage of special accelerators is not uncommon in autonomous driving software because of (i) the extra efforts needed for the understanding of the accelerators and their usage (ii) the benefits from those accelerators are often less satisfying (as we will soon see). 
To make the applications take advantage of DLAs, we revised their implementations. Take YOLOv3 as an example. In the workflow of TensorRT as shown in Figure~\ref{fig:tensorRT}, to build a DLA version of the TensorRT engine for YOLOv3, we set the default execution device to DLA through the invocation of the TensorRT API \path{setDefaultDeviceType(DeviceType::kDLA)}. DLAs do not support all kinds of DNN operators. So to prevent some layers that are unsupported by DLAs (e.g., \emph{LeakyReLU} activation function), it is necessary to make those layers able to fall back to GPUs, which is achieved by adding the invocations of TensorRT API \path{setFlag(BuilderFlag::kGPU_FALLBACK)}. As there are two DLAs on a Jetson AGX Xavier, we can make a particular DLA to execute a kernel through \path{setDLACore(dlaCore)} (dlaCore = 0 or 1) after the deserialization of the TensorRT engine in Figure~\ref{fig:tensorRT}.

Now with DLAs being put into consideration, the scheduling algorithm would need to decide not only the schedules of the nodes on CPUs but also the assignments of the nodes to the three accelerators (one GPU, two DLAs). The design of ROSCH (and HEFT) assumes every node in the DAG can run on any computing unit, which is not the case for the autonomous driving workload where some nodes can run only on CPU, part of some other nodes can run on GPU or DLA. 

To address the issue, we extend the DAG scheduling algorithm in the previous section to make it workable for systems with multiple types of accelerators. The pseudo-code is shown in Algorithm~\ref{fig:exSchAlg}. The design follows two principles (i) being practical (ii) maximizing the quality of the final schedule. An observation is that as our target is a single low-end device, the number of accelerators is very limited (three in the case of Jetson AGX Xavier). Our design hence favors simplicity and result quality over scalability. The basic idea is to enumerate all the possible viable assignments of the task nodes to the accelerators. For each assignment, we instantiate the DAG with the measured performance of the tasks and the communication cost, and then run the scheduling algorithm in the previous section on that DAG to obtain a schedule. In the end, the schedule that gives the best performance is chosen. The assignment corresponding to the DAG gives the final assignments of the task nodes to the accelerators. 


\begin{algorithm}
    \caption{DAG Instantiation Based Scheduling}
    \label{fig:exSchAlg}
    \begin{algorithmic}[1]
        \STATE $D$: the DAG
        \STATE $A=\{a_{i}\}$ : the set of accelerators of one or more kinds
        \STATE $V=\{v_{i}\}$ : the set of task nodes in D
        \STATE $E=\{e_{i}\}$ :  the set of edges in the D
        \STATE $b_{i,j}$: 1 if $v_i$ can run on $a_j$, 0 if $v_i$ cannot run on $a_j$
        \STATE $s$: a valid assignment from $V$ to $A$, that is, \{$b_{i, s(i)}$ = 1 for each $v_i$\}
        \STATE $S=\{s\}$ : the set of valid assignments
        \STATE $Measures$ = \{\}
        \FOR {each $s$ in $S$}
            \STATE $taskPerformance$ $\leftarrow$ measure the performance of tasks under $s$ in the default schedule
            \STATE $d$ = Instantiate $D$ with $taskPerformance$
            \STATE $sch$ = $scheduleAlgorithm(d)$
            \STATE $appPerf$ $\leftarrow$ measure the performance of the application under $sch$
            \STATE $Measures.add(sch, appPerf)$
        \ENDFOR
        \STATE $finalSch$ = $Measures.bestPerf()$
    \end{algorithmic}
\end{algorithm}

\begin{figure*}
    \begin{subfigure}[b]{\textwidth}
        \includegraphics[width=\textwidth]{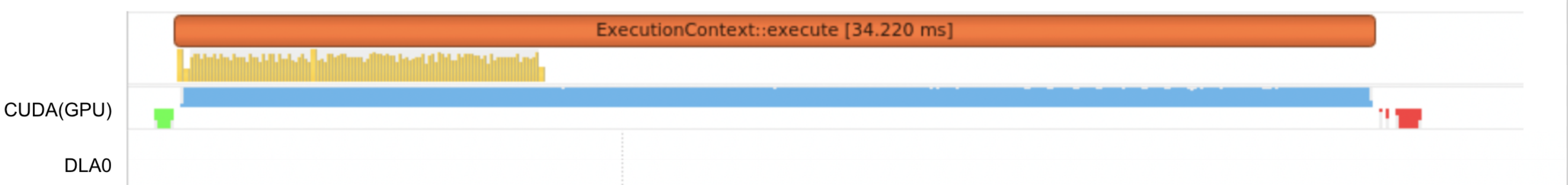}
        \caption{Default execution}\label{subfig:GPU_nsight}
    \end{subfigure}
    \newline
    \begin{subfigure}[b]{\textwidth}
        \includegraphics[width=\textwidth]{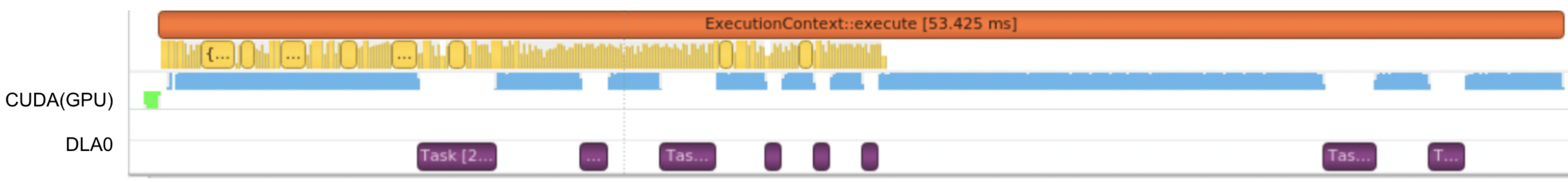}
        \caption{After changes to use DLA}\label{subfig:DLA_nsight}
    \end{subfigure}
    \newline
    \begin{subfigure}[b]{\textwidth}
        \includegraphics[width=\textwidth]{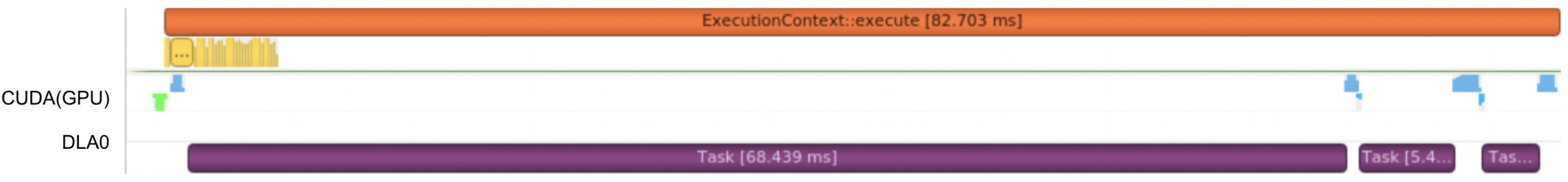}
        \caption{After hardware-aware model customization}\label{subfig:Customize_nsight}
    \end{subfigure}
    \caption{The timeline of the execution of one inference by the YOLOv3-$608\times608$ DNN, collected via NVIDIA Nsight.}
    \label{fig:profiled}
\end{figure*}

\subsection{Performance and Analysis}
\label{sec:performance2}

Segment four in Table~\ref{tab:bigtable} reports the performance after the changes. Compared to the results in Segment 3. where DLAs are not used, the performance becomes worse. The 2D perception becomes faster in two applications (ADy608 and ADs608), but slower in the other applications. The 3D perception becomes slower in all applications.



The reason for the observed performance degradation is that the DLAs are used but only sporadically. DLAs are not as versatile as GPUs. They cannot support some layers in the DNNs. To see the effect of this, using the NVIDIA Nsight profiler, we profile the execution of a YOLOv3-$608\times608$ DNN that has been revised to use DLA. Figure~\ref{fig:profiled} (b) shows the timeline of one inference by the DNN. Compared to the default execution, which does not use the DLA at all, this execution does utilize the DLA, but in only several small-time fragments; most of the time, the execution still happens on the GPU. 


A detailed analysis uncovers the reasons. According to the DLA document~\cite{tensorrtSupport}, the \emph{LeakyReLU} activation function in YOLOv3 is not supported by DLA. In YOLOv3, Convolution, normalization, and activation layers are grouped into one Convolution Block, and the process of feature extraction is composed of multiple Convolution Blocks (There are 57 Convolution Blocks in YOLOv3). If the DLA engine does not support a particular layer, the layer will fall back to GPU execution. In other words, the execution falls back to GPU at least 57 times when making one inference of YOLO-v3. Each of them causes extra memory allocation overhead, and switching overhead on GPU and DLA.

Moreover, the computing engine in TensorRT can hold only up to eight subgraphs. In other words, the falling back to a GPU can happen only up to eight times in an execution. So after the eighth falling back, all remaining layers in the DNN have to run on GPU. As a result, most of the Convolution blocks of YOLOv3 still compete for the GPU resource rather than benefit from the DLAs. 

The six autonomous driving applications include multiple 2D perception DNNs. The switching back and forth between GPUs and DLAs of those DNNs creates interference, especially in the applications with more 2D perception DNNs, which explains the observed increased running time. 

\section{Hardware-Aware Model Customization}
\label{sec:customization}

To address the severe under-utilization of DLAs, the next measure we take is to make changes to the DNN models to make them better fit the restrictions of DLAs. We adhere to three principles: (i) the changes should be minimal; (ii) the changes should cause marginal or no accuracy loss; (iii) the changes can bring significant performance benefits. 

\subsection{Changes}

Based on the analysis described in the previous section, the changes we make are to replace the \emph{LeakyReLU} activation functions in every convolution block in the Yolov3 and Yolov3-SPP models with the standard \emph{Relu} activation function---which is supported by DLAs. One potential concern is the impact of the changes on the accuracy of the DNN. Our experiments show that after re-training (on VOC 2007 dataset~\cite{pascal_voc_2007} which is used in the training of the original v3 and SPP models~\cite{yolov3VOC2007}), the modified models converge to the similar accuracy as the original models do, as Table~\ref{tab:accuracy} shows. 

The results echo the observations made in the previous literature~\cite{xu2015empirical}, the effect of using~\emph{LeakyReLU} over~\emph{Relu} is that the former helps the training converge slightly faster, but the two activation functions do not cause differences in the inference accuracy of Yolov3. 

\begin{table}
    \centering
        \caption{Accuracy before and after the hardware-aware model customization}
    \label{tab:accuracy}
    \begin{tabular}{|c|r|r|}\hline
    Models     & Original & Customized \\\hline
    Yolov3-288    & 77.58\% & 76.73\% \\\hline 
    Yolov3-416    & 80.77\% & 79.75\% \\\hline 
    Yolov3-608    & 78.37\% & 78.06\% \\\hline
    Yolov3-SPP-288     & 78.58\% & 76.89\% \\\hline
    Yolov3-SPP-416     & 81.52\% & 81.15\% \\\hline
    Yolov3-SPP-608     & 78.05\% & 80.07\% \\\hline
    \end{tabular}
\end{table}


\subsection{Performance and Analysis}

Figure~\ref{subfig:Customize_nsight} shows the timeline of the execution of one inference of the modified YOLOv3-$608\times608$. The changes indeed allow the entire DNN model to run on the DLA. It brings two-fold benefits: (i) No falling back to GPUs is needed, and hence the falling back overhead is avoided; (ii) The GPUs are now free of the interference and competition from those models offloaded to DLAs. Because the frequency of DLA is lower than GPU, the inference takes more time than it on GPU. But for an autonomous driving application with many tasks, the effective use of DLA allows the reduction of the load of GPU while still keeping the tasks on DLA meet the deadlines. 

Segment five in Table~\ref{tab:bigtable} reports the performance of the six autonomous driving applications after the change. The same scheduling algorithm as in Section~\ref{sec:migratingToDLA} is used. Now the performance shows significant improvement. All the modules in all of the six applications can now complete their work within the expected latency (As the table caption marks, 10\% over the expected latency is tolerable as such a slack allows real systems to tolerate random fluctuations caused by system noise in real executions.) The applications meet the real-time requirements entirely. Compared to the results in Segment 4, the 3D perception sees about 1.5$\times$ speedups, and the 2D perception sees 2-2.2$\times$ speedups. 


\section{Other Experimented Optimizations}
\label{sec:corun}

In addition to those improvements, we have tried some other optimizations. One of them worth briefly mentioning is an idea we call {\em iterative co-run aware scheduling}. 

\begin{figure}
    \centering
    \includegraphics[width=0.45\textwidth]{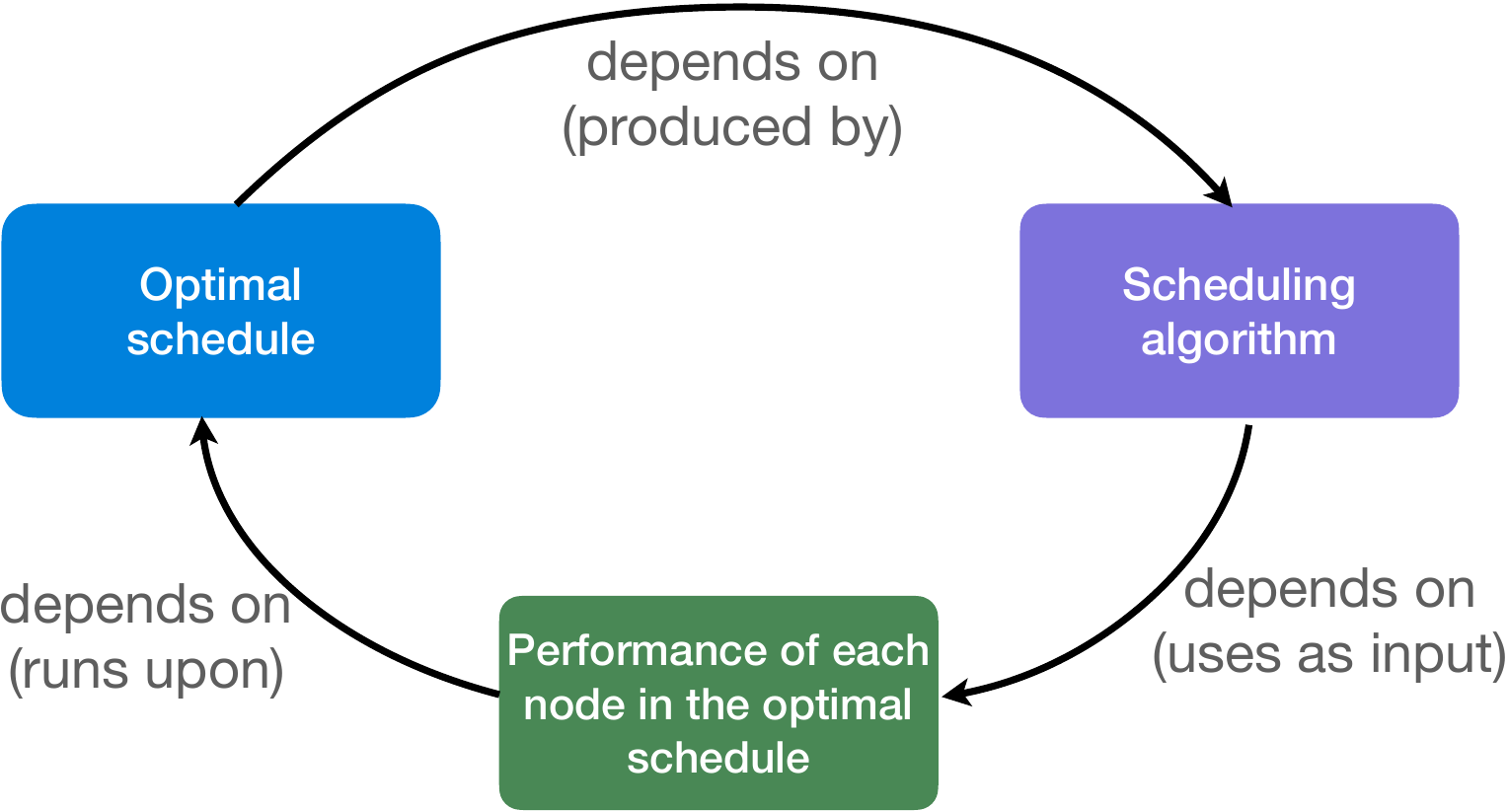}
    \caption{The cyclic dependence in the scheduling problem}\label{fig:chickenEgg}
\end{figure}

\rev{Scheduling algorithms compute the schedules based on the performance of the tasks to schedule. But the performance of the tasks changes as schedule changes. So ideally, the task performance used by all the scheduling algorithms as the basis {\em to find the optimal schedules} should be the performance of the tasks {\em under the optimal schedules}, which forms a cyclic dependence as illustrated in Figure~\ref{fig:chickenEgg}.} 
All scheduling algorithms for autonomous driving have simply used the performance measured in some default schedules as the basis instead. We tried to improve it through an iterative scheme, which resembles the Expectation-Maximization algorithm~\cite{em_algorithm} in machine learning. \rev{It starts with the measured performance of the tasks under a default schedule. It then enters a loop. Each iteration consists of two steps: (i) It applies the scheduling algorithm on the just measured performance of the tasks to calculate a new schedule; (ii) it runs the tasks under that new schedule to remeasure the task performance. It then goes back to step (i), and continues the loop until either the schedule remains unchanged (convergence) or the algorithm times out (i.e., exceeds a threshold).}


The result shows that the method does not yield visible improvement, as Segment six in Table~\ref{tab:bigtable} reports. Our analysis shows that the reason is that the performance of those applications is mostly determined by the performance of the accelerators. Because the scheduling algorithm in Section~\ref{sec:migratingToDLA}, {\em DAG instantiation based scheduling} already considers all the possible placements of a task on each accelerator, and there are only three accelerators, the algorithm already yields good schedules even without the iterative process. For other situations (e.g., many-core CPUs), the algorithm might be more useful; the exploration is beyond the scope of this work.

\section{Power, Synergy, Insights and SCAD Kit}
\label{sec:insights}

Making level-4 autonomous driving possible on a single Jetson AGX Xavier card also gives significant power benefits. Compared to the system mentioned in prior literature~\cite{liu2017computer} that consumes as much as 3000W power, our system consumes only 32W, two orders of magnitude less. Even compared to level-2 autonomous driving systems such as NVIDIA Drive Pegasus (500W), the power consumption is 15$\times$ less. 

A note worth mentioning is that the significant performance improvements in Segment five in Table~\ref{tab:bigtable} come not from an individual technique, but from the synergy of the three measures, JIT priority adjustment, migration to accelerators, and hardware-aware model customization. If we apply only the hardware-aware model customization or the migration to accelerators, the applications would still suffer starvation. If we do not apply the code migration to accelerators, even if the DNN models went through the hardware-aware customization, their code still cannot run on DLAs. 

In Section~\ref{sec:intro}, we have mentioned three main research questions. We summarize the answers offered by this work as follows: 
\begin{itemize}
    \item Answer to RQ1: The current perceived resource needs of autonomous driving are far more than necessary.
    \item Answer to RQ2: Resource under-utilization is the key issue to address to lower the cost of autonomous driving. 
    \item Answer to RQ3: Level-4 autonomous driving can achieve real-time performance on a single off-the-shelf card for as little as \$699. To achieve that, it needs a synergy of just-in-time affinity and priority adjustment, appropriate code migration to accelerators, and hardware-aware model customization. 
\end{itemize}

A side product from this study is the SCAD kit, a package that consists of not only the six ADApp applications, but also a flexible generator of autonomous driving applications customized to fully utilize DLAs, builtin just-in-time priority adjustment, and several options of scheduling algorithms (non-iterative, iterative scheduling). The kit allows easy deployment of autonomous driving on a Jetson card, as well as the generation of various autonomous driving DAGs for experiments and benchmarking through its reconfiguration scheme. Nodes of various tasks can be easily added or removed from the DAG by revising a configuration file. By addressing the many limitations of existing frameworks for autonomous driving research, SCAD offers a vehicle for the community to more quickly advance the research in this field. It will be released to the public after the publishing of this work.

\section{Implications}
\label{sec:discuss}

The results from this work are significant. Before this work, no prior work has ever shown that level-4 autonomous driving is possible to achieve real-time performance on a single off-the-shelf low-end car-like Jetson AGX Xavier. The cost, \$699, is an order of magnitude less than those used in the current industry (e.g., the \$30k system described in previous literature~\cite{liu2017computer}). 

Note that besides the real-time performance requirement, a level-4 autonomous driving system must have enough redundancy, reliability and security to ensure safety. There are two approaches to adding redundancy, with both considered in our product design. A heterogeneous approach is to have multiple channels to create their own independent, and diverse perceptions of the world \cite{truered} for cross-checks. A homogeneous approach is to run the same workload on two identical devices with redundant power supplies equipped on each device~\cite{talpes2020compute}. What this paper focuses on is performance of the core AI component. Being able to run the entire application on such a low-end card prepares a good basis: Even if the addition redundancy doubles or triples the cost, the cost would still be an order of magnitude lower than state of the art.


Besides pointing out a promising path for the industry to drastically reduce the cost and power of autonomous driving systems, the overturning of the common perceptions by this work also suggests some new research opportunities.  Some examples are as follows: 

\begin{itemize}
\item Architecture design: As the cost and power consumption drop dramatically, it would be valuable to reexamine the design of the entire autonomous driving architecture in terms of the budget allocation for various components, reinforcement of security or reliability, and so on.
\item Software design: The changed assumptions on the cost and computing resource suggests the need for rethinking the design of the autonomous driving software, such as the complexities and structures of adoptable DNNs, the inter-component communication, and synchronizations, scheduling, and so on.
\item Optimizations: There have been lots of research on the optimization of certain points in autonomous driving. They may be worth reexamination. For instance, as everything now runs on a single card, inter-card communication becomes less important, but how to effectively improve the reliability of the low-end device becomes more important. The many optimizations proposed before may work differently in this single-card setting. The research to schedule each layer of a DNN on GPUs to strike an accuracy-power-speed tradeoff~\cite{bateni2020neuos}, for example, may now need to consider the (core, data path, and memory) contentions from other DNN models running on both GPUs and other accelerators (e.g., DLAs). 
\end{itemize}



\section{Related Work}



To cope with the complexity in level-4 driving, many autonomous machine companies resort to \textit{ad hoc} solutions to ensure on-time autonomous machine product release. These \textit{ad hoc} solutions are often product-specific, evolve through trial and error, and are hard to generalize to other autonomous machine designs, hence leading to high re-engineering costs for each product. For instance, in the solution presented in \cite{yu2020building}, the computing platform consists of a Xilinx Zynq UltraScale+ FPGA board and an on-vehicle PC machine equipped with an Intel Coffee Lake CPU and an Nvidia GTX 1060 GPU. Many manual efforts are spent in customizing the software deployment, scheduling, and resource management to such ad hoc solutions~\cite{yu2020building,liu2020autonomous}.

LoPECS~\cite{tang2020lopecs} is an effort to make autonomous driving software runnable on an embedded low-end device. The autonomous driving software is, however, much simpler than level-4 autonomous driving applications. Its vision sensing, for instance,  consists of only one camera, and its perception module consists of only one CNN model; there are no 3D perception, other 2D perception models, or perception fusions. Even with that simplicity, it still needs to offload extensive computations to the cloud, the uncertain delays incurred by which add risks to the safety of the system. Due to the simplicity of the problem in the prior work, none of the complexities encountered in this work---such as the priority-related starvation, the use of non-GPU accelerators, the scheduling for multi-type accelerators---manifests in their work. 


Several studies proposed multi-DNN schedulers on heterogeneous SoC. 
ApNet \cite{bateni2018apnet} applies approximation approaches to each layer of the DNN network and makes a trade-off between accuracy and latency. PredJoule \cite{bateni2018predjoule} optimizes energy for running DNN workloads. It adjusts power configuration based on the latency of workloads. NeuOS  \cite{bateni2020neuos} provides a system for running multiple DNN workloads for autonomous systems. It tries to balance three dimensions of optimization: latency, accuracy, and power consumption. The idea of the NeuOS is to schedule multi-DNN layer by layer. In each layer boundary (layer end execution), the scheduler will decide the power config of the whole system and the accuracy of the next layer based on assigned policy (min energy, max accuracy, or balanced both) and the deadline of each network. 


Even though those studies provide useful insights at certain focused research points, the practical situation is more complicated in the autonomous driving system. It has various modules such as sensing, localization, and so on. Each module not only contains multiple DNNs to support, but also depends on each other. And the entire applications consists of tasks of varied kinds, some runnable only on CPU, some involving CPU parts and accelerator parts, and the parts for accelerators may use only GPU or both GPU and other accelerators (e.g., DLA). 

Many recent studies present schedulers for running multiple DNN workloads on the cluster. These cluster schedulers~\cite{xiao2018gandiva, mahajan2019themis, gu2019tiresias, le2020allox, narayanan2020heterogeneity} attempt to make an optimal decision to allocate diverse resources to many requests of DNN workloads. They have several optimizing objectives, such as fairness and throughput, to design-related scheduling policies. They adopt deployment optimizations such as space sharing in Gandiva and placement sensitivity in Themis and Tiresias to increase resource utilization. The settings, constraints, and considerations on embedded systems are all different. 

\section{Conclusion}

This paper has presented an experience that manages to make level-4 autonomous driving applications achieve real-time performance on a single low-end card at a cost an order of magnitude lower than the devices used in today's autonomous driving industry. It achieves it by addressing three major deficits in the current practices through several practical solutions, which include {\em just-in-time affinity and priority adjustment}, {\em model migrations to all types of accelerators}, {\em DAG instantiation based scheduling} and {\em hardware-aware model customization}. 

This work points out a promising path for the industry to drastically lower the cost and power consumption of autonomous driving. By overturning the common assumptions on the required computing resource by autonomous driving, the work suggests rethinking the current architectures, software, and optimization for autonomous driving, and opens up many potential research opportunities in various dimensions.

\section*{Acknowledgement}

This material is based upon work supported by the National Science
Foundation (NSF) under Grants CCF-1703487 and CCF-2028850. Any
opinions, findings, and conclusions or recommendations expressed in
this material are those of the authors and do not necessarily reflect
the views of NSF.

\bibliographystyle{IEEEtranS}
\bibliography{refs}
\balance
\end{document}